\documentclass[a4paper]{article}

\usepackage{INTERSPEECH2021}
\usepackage{lipsum}
\usepackage{todonotes}
\usepackage{lineno}
\usepackage{multirow}
\usepackage{url}
\usepackage{soul}

\newcommand{\BERT}[1]{BERT\textsubscript{#1}}

\title{Disfluency Detection with Unlabeled Data and Small BERT Models}
\name{Johann C. Rocholl, Vicky Zayats, Daniel D. Walker, \\
      Noah B. Murad, Aaron Schneider, Daniel J. Liebling}
\address{Google Research, USA}
\email{\{jcrocholl, vzayats, danwalkeriv, noahmurad, aaronschneider, dliebling\}@google.com}

\begin{document}

\maketitle

\begin{abstract}
Disfluency detection models now approach high accuracy on English text. However, little exploration has been done in improving the size and inference time of the model. At the same time, Automatic Speech Recognition (ASR) models are moving from server-side inference to local, on-device inference. Supporting models in the transcription pipeline (like disfluency detection) must follow suit. In this work we concentrate on the disfluency detection task, focusing on small, fast, on-device models based on the BERT architecture. We demonstrate it is possible to train disfluency detection models as small as 1.3 MiB, while retaining high performance. We build on previous work that showed the benefit of data augmentation approaches such as self-training. Then, we evaluate the effect of domain mismatch between conversational and written text on model performance. We find that domain adaptation and data augmentation strategies have a more pronounced effect on these smaller models, as compared to conventional BERT models.
\end{abstract}


\section{Introduction}
\label{sec:intro}

On-device AI models have many benefits, including lower inference time, privacy preservation, and the ability to operate without an internet connection. As a result, many apps that transcribe spontaneous speech are moving away from server-side to on-device Automatic Speech Recognition (ASR) pipelines. ASR output may contain noise which affects the readability of such transcripts. Sources of noise include ASR errors, incorrect sentence boundaries, filler words, and disfluencies (filled pauses and self-repairs). These errors can reduce the performance of downstream tasks such as natural language understanding (NLU) or machine translation. Disfluency detection is one way to improve the readability of ASR transcripts, and here we focus on small and fast models based on the BERT architecture \cite{devlin-etal-2019-bert} that are suitable for use in on-device ASR pipelines.

Disfluencies are irregularities that are an integral part of spontaneous speech and include self-repairs, repetitions, restarts, and filled pauses \cite{Schegloff1977}. Following Shriberg et al. \cite{shriberg1997prosody}, the disfluency annotation includes the \textit{reparandum} (the material that the speaker intends to delete), \textit{interruption point} ($+$), optional \textit{interregnum} which includes filled pauses and discourse markers (e.g. ``uh,'' ``um,'' ``you know,'', ``I mean''), and an optional \textit{repair} --- the material that semantically replaces the reparandum. Below are a few examples of disfluent phrases:

{\small
\begin{verbatim}
 [ it's + { uh } it's ] almost... 
 [ was it, + { I mean, } did you ] put...
 [ By + ] it was attached to...
 \end{verbatim}
}

\noindent
Recent developments of smaller on-device BERT-based models (e.g. MobileBERT \cite{sun2020mobilebert} and Small-vocab BERT \cite{zhao-etal-2021-extremely}) have adapted this architecture for use in resource-constrained contexts including mobile devices. These models are usually pretrained using written text like Wikipedia articles and books, contributing to a domain mismatch with spontaneous speech. Here, we demonstrate that it is possible to train disfluency detection models that are small enough to run in real time on-device, while retaining a high level of accuracy. We build on previous work that showed the benefit of data augmentation approaches such as adding simulated disfluency data and self-training to show how self-training and model pretraining on conversational data can benefit on-device models.

The contributions of this paper are as follows. First, to the best of our knowledge, this is the first work that explores the capabilities of small on-device disfluency detection models, showing that it is possible to achieve only slight degradation in performance on a disfluency detection task with a model as small as 1.3 MiB. This represents a model size reduction by two orders of magnitude compared to a state-of-the-art \BERT{BASE} model and inference latency reduction by a factor of 8. We also show the importance of pretraining and the effect of domain mismatch between conversational and written text on model performance. In particular, we find that self-training has a more pronounced effect on these smaller models, as compared to conventional BERT models, while pretraining on Reddit improved the performance of a large \BERT{BASE} model. 
\section{Related Work}

Prior to BERT, LSTM-based models and their variants were a popular choice for disfluency detection task. Those approaches included more traditional usage of LSTMs \cite{DBLP:journals/corr/ZayatsOH16, wang2016neural}, LSTM variants that explicitly exploited similarity between the reparandum and repair of disfluencies \cite{lou2018disfluency,zayats2018robust, wang2017transition}, and noisy-channel model \cite{lou2017disfluency}. After the development of BERT, larger BERT-based models showed significant improvement over prior work on disfluency detection task \cite{bach2019noisy, jamshid-lou-johnson-2020-improving}.

Prior studies on disfluency detection showed the importance of data augmentation approaches, predominantly due to the relatively small size of Switchboard corpus \cite{Switchboard} that is used as the main source of annotated data. Both Wang et al. \cite{wang-etal-2018-semi} and Bach et al. \cite{bach2019noisy} showed improvement by simulating and inserting disfluencies in fluent transcripts and using those transcripts as an additional source of data. Similarly, self-training proposed by Jamshid Lou et al. \cite{jamshid-lou-johnson-2020-improving} augments Switchboard corpus by using a larger Fisher corpus \cite{fisher} automatically labeled by an existing disfluency classifier model. While some prior studies incorporate prosody \cite{ferguson2015disfluency,Zayats2019} or use the acoustic signal as an input \cite{kourkounakis2020fluentnet,lou2020end}, in this work we use manually transcribed text only, leaving the usage of acoustic signals and ASR transcripts for future work.
\section{Methods}

In this work we explore BERT and some of its more compact variants fine-tuned for the disfluency detection task. Following previous work on disfluency detection with BERT \cite{bach2019noisy,jamshid-lou-johnson-2020-improving} our biggest model that is used for pretraining and self-training is the \BERT{BASE} \cite{devlin-etal-2019-bert} model. Below we provide an overview of some smaller and faster models that are specifically targeted for on-device usage.

\subsection{Size Reduction}
\label{sec:methods}
Recently a number of BERT distillation approaches have been proposed that allow significant size and latency reduction. Most of those models follow the student-teacher paradigm where a large BERT model plays the role of a teacher, while a compact student model (with reduced number of layers, hidden states, number of heads and/or vocabulary size) learns the teacher behaviour using various distillation techniques. In this work we tried the following pretrained distilled BERT models:

\begin{description}
    \item[DistilBERT \cite{sanh2019distilbert}] In addition to a masked language model (MLM) objective, DistilBERT uses knowledge distillation approach \cite{hinton2015distilling}, where a student is trained to recover the predictions of a teacher.
    
    \item[MobileBERT \cite{sun2020mobilebert}] In addition to layer-wise knowledge distillation, MobileBERT uses bottleneck structures that allow a model projection from the teacher to the student.
    
    \item[TinyBERT \cite{jiao2019tinybert}] learns a direct mapping for embedding, hidden state, attention, and prediction layers in order to transfer model knowledge from teacher to student. 
    
    \item[PD-BERT \cite{turc2019well}] introduced a set of very small models that are first pretrained using MLM objective, and only then use knowledge distillation similar to DistilBERT.
    
    \item[Small-vocab BERT \cite{zhao-etal-2021-extremely}] uses a mixed-vocab training approach to train very small models with a smaller WordPiece vocabulary (5k vs. the usual 30k) because for small models, the token embedding layer accounts for a significant percentage of overall model size.
\end{description}

While some of these approaches propose to fine-tune the teacher model on the task before distillation happens, in our current work for all the above mentioned models we start with an already distilled student model and fine-tune it on disfluency detection task.

\subsection{Domain Adaptation}
Commonly available pretrained BERT models were trained using text from Wikipedia and Books. While this exposes the model to a large set of topics, the linguistic styles of these corpora do not necessarily match the more conversational and disorganized style of spontaneous speech and dialogue. In this paper we experiment with self-training and pretraining in order to mitigate domain shift caused by mismatch of the styles.

\subsubsection{Self-training}
We follow Jamshid Lou et al. \cite{jamshid-lou-johnson-2020-improving} by using self-training with the Fisher corpus. Specifically, first we train (fine-tune) a full-size \BERT{BASE} model on Switchboard corpus and then use that model to add predicted disfluency labels to the Fisher corpus. After that, we use this automatically annotated Fisher corpus, also referred here as ``silver'' data, as an additional source of training data for further fine-tuning of the \BERT{BASE} or smaller student model.

\subsubsection{Pretraining}
 We hypothesized that, just like self-training on additional conversational text can help the model generalize better, starting with a model that is pretrained on more conversational data will improve the model's ability to create good spontaneous speech representation, leading to better performance in the disfluency classification task. In this work we started with Wiki/Books-pretrained \BERT{BASE} and one of the Small-vocab BERT models and continued the pretraining separately on two datasets that were chosen to more closely match casual and spontaneous speech: Reddit comments and transcribed human speech from the Fisher dataset \cite{fisher} (see Section~\ref{sec:data}).

\section{Experiments}

\begin{table}[t]
  \caption{Datasets statistics.}
  \label{tab:data_stats}
  \centering
  \begin{tabular}{|l|c|c|c|c|c|} \hline
    \textbf{Dataset} &
    \rotatebox{90}{\textbf{Sentences}} &
    \rotatebox{90}{\textbf{Docs}} &
    \rotatebox{90}{\textbf{Words}} &
    \rotatebox{90}{\parbox{1.4cm}{\textbf{Disfluent Sentences}}} &
    \rotatebox{90}{\parbox{1.4cm}{\textbf{Disfluent Spans}}}  \\ \hline
    Switchboard  & 77k   & 1k   & 1M     & 25k  & 34k    \\ \hline
    Fisher       & 1.3M  & 12k  & 20M    & 492k & 676k   \\ \hline
    Reddit       & 314M  & 135M & 3B    & N/A   & N/A     \\ \hline
  \end{tabular}
  \vspace{-0.4cm}
\end{table}

\begin{table*}[t]
  \caption{Model size vs performance on Switchboard dev dataset. Asterisk indicates models using Wiki/books checkpoints available via Huggingface Transformers \cite{wolf-etal-2020-transformers}. For the rest of the models we used compatible initial checkpoints pretrained on the same Wiki/books dataset that have not yet been released. Bold font highlights the smallest models with good performance.}
  \label{tab:size_reduction}
  \centering
  \begin{tabular}{|l|c|c|c|c|c|c|c|c|c||c|c|c|} \hline
    \textbf{Model} &
    \rotatebox{90}{\textbf{Layers}} &
    \rotatebox{90}{\textbf{Hidden Size}} &
    \rotatebox{90}{\textbf{Heads}} &
    \rotatebox{90}{\textbf{Params}} &
    \rotatebox{90}{\textbf{Size (MiB)}} & 
    \rotatebox{90}{\textbf{Latency (ms)~}} &
    \rotatebox{90}{\textbf{Precision}} &
    \rotatebox{90}{\textbf{Recall}} &
    \textbf{F1} &
    \rotatebox{90}{\textbf{Precision}} &
    \rotatebox{90}{\textbf{Recall}} &
    \textbf{F1} \\ \hline
    \BERT{BASE}   \cite{devlin-etal-2019-bert}
    & 12 & 768 & 12 & 108.9M & 416 & 72 & 92.9 & 90.2 & 91.5 & 93.2 & 90.5 & 91.8 \\ \hline
    DistilBERT* \cite{sanh2019distilbert}
    & 6 & 768 & 12 & 66.4M & 253 & 42 & 93.2 & 87.7 & 90.4 & 93.8 & 88.9 & 91.3 \\ \hline
    \multirow{2}{*}{TinyBERT* \cite{jiao2019tinybert}}
    & 6 & 768 & 12 & 66.4M & 253 & 42 & 93.6 & 87.0 & 90.2 & 93.5 & 88.8 & 91.1 \\ \cline{2-13}
    & 4 & 312 & 12 & 14.3M &  54 & 13 & 90.6 & 84.1 & 87.2 & 91.7 & 87.5 & 89.5 \\ \hline
    MobileBERT* \cite{sun2020mobilebert}
    & 24 & 128 & 4 & 24.6M &  95 & 41 & 92.4 & 89.7 & 91.0 & 92.7 & 89.9 & 91.3 \\ \hline
    \multirow{4}{*}{Small-vocab BERT \cite{zhao-etal-2021-extremely}}
    & 24 & 192 & 3 & 11.7M & 45 & 32 & 90.9 & 87.8 & 89.3 & 91.5 & 89.0 & 90.2 \\ \cline{2-13}
    &  6 & 256 & 8 & 4.6M &  18 &  12 & 91.9 & 85.3 & 88.5 & 92.8 & 87.5 & 90.1 \\ \cline{2-13}
    & \textbf{12} & \textbf{128} & 4 & 3.1M &  \textbf{12} & 15 & 91.4 & 86.4 & 88.8 & 
    \textbf{92.5} & \textbf{88.2} & \textbf{90.3} \\ \cline{2-13}
    & \textbf{6} & \textbf{96} & 4 & 1.2M & \textbf{4.7} & 9 & 90.3 & 79.1 & 84.3 & 
    \textbf{92.5} & \textbf{85.5} & \textbf{88.9} \\ \hline
    \multirow{2}{*}{PD-BERT* \cite{turc2019well}}
    &  4 & 128 & 2 & 4.8M &  18 &   7 & 91.9 & 77.9 & 84.3 & 92.4 & 85.0 & 88.5 \\ \cline{2-13} 
    &  2 & 128 & 2 & 4.4M &  17 &   5 & 89.6 & 72.1 & 79.9 & 92.8 & 79.9 & 85.9 \\ \hline
    \multicolumn{7}{c}{}
    & \multicolumn{3}{|c||}{\textbf{baseline}}
    & \multicolumn{3}{|c|}{\textbf{self-trained}} \\ \cline{8-13}
  \end{tabular}
  \vspace{-0.3cm}
\end{table*}

\subsection{Data and Tokenization}
\label{sec:data}
For training, we use transcripts from Switchboard corpus of English conversational dialogue \cite{Switchboard}. This corpus is widely used in disfluency detection research because it has gold labels for disfluent spans of text. We use the established train/dev/test splits \cite{charniak2001edit}. We removed commas and filled pauses (``uh", ``huh", ``uh-huh", ``um") because they are not found in most of the ASR output that we want to annotate with our models.

In most of our experiments, unless mentioned otherwise, we report metrics on the task of labeling both the reparandum and interregnum (e.g. ``you know", ``well", ``I mean") of disfluencies. This is in contrast to the majority of prior studies that typically focus on the recognition of the reparandum only. Our motivation here is to identify spans that should be elided in order to improve the overall readability of a transcript. For the purpose of comparison to prior work, we also include results for our best-performing model configurations retrained using the standard reparandum-only task (see Section~\ref{sec:comparison}).

In this work we experiment with using the Fisher dataset \cite{fisher} for both pretraining and self-training. In order to reduce fragmentation, we preprocess Fisher by combining utterances across interruptions and combining short contiguous utterances from the same speaker. For the next sentence prediction task used during model pretraining, we sorted the processed utterances by the timestamp of the first component utterance.

In addition to the Fisher corpus, we experiment with using Reddit comments for pretraining our models. The main benefit of using Reddit is that it is a large corpus of casual/conversational discussions. Since it is text-based, it doesn't contain disfluencies like the spontaneous speech recorded in Fisher, but it is still much more conversational than most text from Wikipedia. We used the 2019 portion of the Pushshift dataset \cite{baumgartner2020pushshift}. For the next sentence prediction pretraining task, each comment was treated as a separate document and segmented into sentences using NLTK's\footnote{\url{https://www.nltk.org}} \texttt{sent\char`_tokenize}.

Table \ref{tab:data_stats} shows summary stats on the Switchboard training, Fisher, and Reddit sets, including the numbers of total sentences, ``documents" (comments/conversations), words, sentences with a tagged disfluency, and total spans tagged as disfluent in each. For the last two we use labels from \BERT{BASE} model for Fisher, and human-annotations for Switchboard.

\subsection{Metrics}
\label{section:metrics}
Our main metric for model performance is token-level F1 score (the harmonic mean of token-level precision and recall). We report model sizes in number of parameters and/or mebibytes (1 MiB = $1024^2$ bytes). Latency is measured in milliseconds for the median time that it takes to perform batch inference on 8 example sentences on GPU.

\subsection{Hyper-parameter Tuning}
\label{section:hyperparameters}
We optimize the learning rate (around 2e-4), training batch size (from 32 to 512), number of training epochs (up to 55 for small models), and (for self-training) the percentage of ``silver" data included in each training batch (around 70\%). We used a black-box hyper-parameter optimization system \cite{vizier}. For most experiments we ran 64 trials total, with 8 evaluations in parallel. Each individual trial (one set of hyper-parameters) ran on a single NVIDIA P100 GPU for about 30 minutes to 2 hours. All results shown are parameterizations that yielded the best F1 score on the Switchboard dev set.

\begin{figure}[b!]
    \vspace{-1mm}
    \scalebox{0.8}{\input{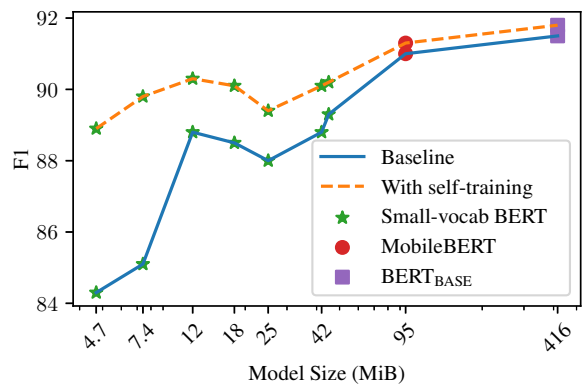}}
    \vspace{-4mm}
    \caption{F1 scores for model configurations with and without self-training, shows performance correlated with model size, self-training helps more for small models. Not all models shown, though similar patterns exist for other model types.}
    \label{fig:size_vs_f1}
    \vspace{-0.2cm}
\end{figure}

\subsection{Results}
First, we compare the model performance for BERT and its smaller variants mentioned in Section~\ref{sec:methods} of different sizes that were pretrained on Wikipedia and books corpus, results are presented in Table~\ref{tab:size_reduction}. As expected, model performance generally declines with the size of the model. While comparing different models, we found that among the tiniest of the models that are capable to fit on-device, Small-vocab BERT has the best performance. 
In addition to evaluating various models using Switchboard corpus, we also experiment with incorporating self-training approach using ``silver" Fisher data. The results for self-training experiments are also presented in Table~\ref{tab:size_reduction}. While performing self-training for models of different sizes we have noticed an interesting trend --- the amount to which self-training improves model performance appears to increase as the model size decreases, making self-training especially important for smaller models. This trend can be seen more easily in Figure~\ref{fig:size_vs_f1} on \BERT{BASE}, MobileBERT, and Small-vocab BERT models, and similar trends were observed with PD-BERT models. By looking at the predictions on the dev set we did not notice much difference in prediction quality between \BERT{BASE} and Small-vocab BERT besides the slightly elevated number of errors associated with false-positive repetition detection in a Small-vocab BERT.

\begin{table}[b]
  \caption{Results of 8-bit quantization using TensorFlow Lite. Model size (MiB) about 28\% of floating point model, latency (ms) for single-sentence inference on Android Pixel 3a, performance degradation on Switchboard dev set $\leq$ 0.2 points.}
  \label{tab:quant}
  \centering
  \begin{tabular}{|l|l|c|c|c|c||c|} \hline
    \multicolumn{2}{|l|}{\textbf{Model}} & \textbf{Size} & \textbf{Lat} & \textbf{Prec} & \textbf{Rec} & \textbf{F1} \\ \hline \hline
    \multicolumn{2}{|l|}{MobileBERT} & 25 & 162 & 92.6 & 89.9 & 91.2 \\ \hline
    Small- & 12$\times$128          & 3.3 & 32 & 92.3 & 88.0 & 90.1 \\ \cline{2-7}
    vocab & 6$\times$96             & 1.3 & 11 & 92.4 & 85.4 & 88.8 \\ \hline  
  \end{tabular}
  \vspace{-0.3cm}
\end{table}

\subsection{Effect of Pretraining}
In order to measure the impact of domain mismatch between the pretraining data and the target data (spontaneous speech) on final model performance, we experiment with BERT pretraining using the Fisher and Reddit corpora for a large \BERT{BASE} configuration and Small-vocab BERT 12$\times$128. Following Devlin et al. \cite{devlin-etal-2019-bert} we restrict pretraining to 1M steps with an initial learning rate of 0.0001 which decreased in equal increments at each step to zero. In order to identify the best stopping criteria, we evaluate model performance on disfluency detection task every 200K steps. For Fisher the optimal checkpoint was found after pretraining for 200K-400K steps, while for Reddit the optimal stopping point was at 600K-1M steps, which can be explained by the large size disparity between Fisher and Reddit. 

\begin{figure}[ht]
    \scalebox{0.8}{\input{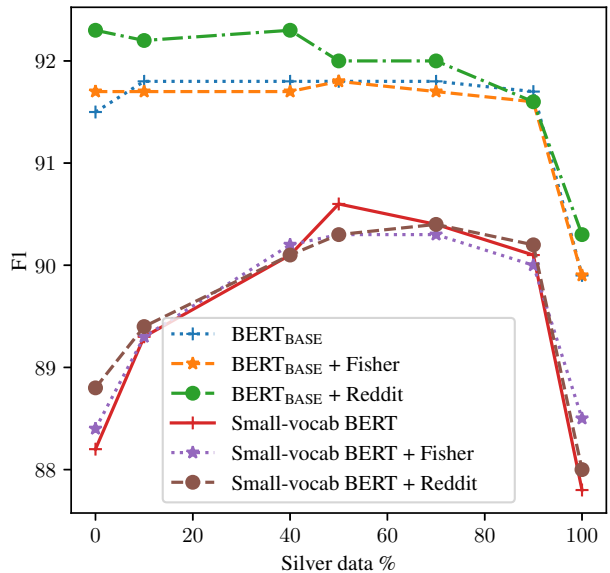}}
    \vspace{-6mm}
    \caption{Self-training silver data percentage v.s. F1 for the \BERT{BASE} and Small-vocab BERT 12$\times$128 configurations, with and without pretraining on Reddit or Fisher.}
    \label{fig:silver_pct}
\end{figure}

In order to understand the impact of pretraining, we ran experiments combining pretraining and self-training in various configurations. For all of our 6 settings, we experiment with including different percentages of silver data and plot F1 scores for \BERT{BASE} and Small-vocab BERT 12$\times$128 in Figure~\ref{fig:silver_pct}. We used a derivative free optimization package to optimize the training parameters (batch size, learning rate, number of iterations, etc.) and chose the "best" result for each model configuration at each silver data percentage which is likely why this figure differs somewhat from that in \cite{lou2018disfluency}.
From those trends we notice that pretraining on Reddit benefits the larger \BERT{BASE} model, achieving F1 score 92.3, while not necessarily helping smaller model (F1 score 90.4).

Most of the self-training results presented here use the same Fisher corpus, with disfluency labels predicted by a \BERT{BASE} model fine-tuned on Switchboard only. As an additional experiment, we used our best \BERT{BASE} model from Table~\ref{tab:no_interregnum} (pretrained on Reddit, self-trained on Fisher, and hyper-parameter tuned for best performance on Switchboard dev set, F1 score 90.9 on test set) to predict better labels on Fisher corpus. When using this improved dataset for self-training the Small-vocab BERT 6$\times$96 model, we found no significant performance improvement on Switchboard test set, presumably because this tiny model was already pretty close to optimal performance for its size. But for the 12$\times$128 model, this iterative approach improved precision, recall and F1 score by about 0.4 points, which is similar to the 0.5 point improvement for \BERT{BASE} + Reddit + self-training.

\subsection{Comparison to Other Work}
\label{sec:comparison}
Finally, we compare our best performing models against previously published results using Switchboard test set in Table~\ref{tab:no_interregnum}. To make our findings comparable, for this set of experiments we retrained and evaluated the models with labels associated with reparandum only, without explicitly marking the interregnum. Unfortunately we were not able to find published size or latency numbers for all of these results.

\begin{table}[h]
  \caption{Model comparison on Switchboard test set without including interregnum in the label.}
  \label{tab:no_interregnum}
  \centering
  \begin{tabular}{|l|l|c|c||c|} \hline
    \textbf{Arch} & \textbf{Model} & \textbf{Prec} & \textbf{Rec} & \textbf{F1}\\ \hline \hline
    CRF & Ferguson et al. \cite{ferguson2015disfluency} & 90.1 & 80.0 & 84.8 \\ \hline
    \multirow{3}{*}{LSTM}
    & Zayats et al. (2016) \cite{DBLP:journals/corr/ZayatsOH16} & 91.8 & 80.6 & 85.9 \\ \cline{2-5}
    & Jamshid Lou et al. \cite{lou2018disfluency}               & 89.5 & 80.0 & 84.5 \\ \cline{2-5}
    & Zayats et al. (2018) \cite{zayats2018robust}              & - & - & 86.7 \\ \hline 
    \multirow{5}{*}{BERT}
    & Bach et al. (2019) \cite{bach2019noisy}                      & 94.7 & 89.8 & 92.2 \\ \cline{2-5}
    & Jamshid Lou et al. \cite{jamshid-lou-johnson-2020-improving} & 86.7 & 91.9 & 89.2 \\ \cline{3-5}
    & + ensemble of 4 models                                       & 87.5 & 93.8 & 90.6 \\ \cline{2-5}
    & \BERT{BASE} (416 MiB)         & 92.6 & 88.4 & 90.4 \\ \cline{3-5}
    & + Reddit + self-trained       & 93.1 & 88.9 & 90.9 \\ \hline
    \multirow{4}{*}{\parbox{5mm}{Small-vocab BERT}}
    & 12$\times$128 self-trained    & 92.0 & 87.9 & 89.9 \\ \cline{3-5}
    & + quantized (3.3 MiB)         & 91.9 & 87.8 & 89.8 \\ \cline{2-5}
    & 6$\times$96 self-trained      & 91.3 & 85.9 & 88.5 \\ \cline{3-5}
    & + quantized (1.3 MiB)         & 91.1 & 85.8 & 88.4 \\ \hline
  \end{tabular}
  \vspace{-0.3cm}
\end{table}

\section{Conclusions and Future Work}

Self-training can be used to transfer knowledge from a large teacher model to a much smaller student. We present results for several experiments with different combinations of the number of layers, hidden size, attention heads, vocabulary size, and distillation methods. Additional pre-training on conversational speech corpora helps improve disfluency detection performance compared to pretraining only on Wikipedia+Books.
Quantization can reduce model size even more.
By combining these techniques, we reduced model size by 99\% and inference latency by 80\% while maintaining competitive performance.

While the focus of this paper is on investigating simple semi-supervised learning techniques for task-agnostic distilled models, we leave it to future work to understand if we can further improve model performance by using semi-supervised techniques before or during model distillation.
\section{Acknowledgements}

The authors would like to thank everyone who provided feedback on drafts of this paper, especially Shyam Upadhyay, Raghav Gupta, and Dirk Padfield. Thanks also to Max Gubin for supplying the processing scripts for the Reddit data.

\bibliographystyle{IEEEtran}
\bibliography{mybib}

\end{document}